\pdfoutput=1
\documentclass{article}

\usepackage{arxiv}

\usepackage[utf8]{inputenc} 
\usepackage{cite}
\usepackage{amsmath,amssymb,amsfonts}
\usepackage{algorithmic}
\usepackage{graphicx}
\usepackage{textcomp}
\usepackage{hyperref}
\usepackage{xcolor}
\usepackage{caption}
\usepackage{listings}
\usepackage{dirtree}
\usepackage{mdframed}

\hypersetup{
    colorlinks=true,
    linkcolor=black,  
    citecolor=black,  
    urlcolor=blue     
}
\def\BibTeX{{\rm B\kern-.05em{\sc i\kern-.025em b}\kern-.08em
    T\kern-.1667em\lower.7ex\hbox{E}\kern-.125emX}}

\title{UncertaintyPlayground: A Fast and Simplified Python Library for Uncertainty Estimation}

\author{Ilia Azizi\\
University of Lausanne\\
Lausanne, Switzerland \\
\texttt{ilia.azizi@unil.ch}\\
}

\begin{document}
\maketitle

\begin{abstract}
This paper introduces \textit{UncertaintyPlayground}, a Python library built on PyTorch and GPyTorch for uncertainty estimation in supervised learning tasks. The library offers fast training for Gaussian and multi-modal outcome distributions through Sparse and Variational Gaussian Process Regressions (SVGPRs) for normally distributed outcomes and Mixed Density Networks (MDN) for mixed distributions. In addition to model training with various hyperparameters, UncertaintyPlayground can visualize the prediction intervals of one or more instances. Due to using tensor operations, the library can be trained both on CPU and GPU and offers various PyTorch-specific techniques for speed optimization. The library contains unit tests for each module and ensures multi-platform continuous integration with GitHub Workflows (online integration) and Tox (local integration). Finally, the code is documented with Google-style docstrings and offers a documentation website created with MkDocs and MkDocStrings. 
\end{abstract}


\section{Introduction}
Uncertainty estimation is a critical aspect of machine learning, providing insights into the reliability of predictions. While efficient libraries such as PyTorch \cite{paszke2019pytorch} and GPyTorch \cite{gardner2018gpytorch} offer great flexibility, they are designed to be relatively low-level and, deliberately, do not offer process abstraction. This paper introduces UncertaintyPlayground\footnote{\url{https://unco3892.github.io/UncertaintyPlayground/}}, a Python library that facilitates uncertainty estimation in supervised learning tasks, offering a user-friendly yet performant interface to powerful, parallelized uncertainty estimation techniques.

The advent of Neural Networks (NNs) and Graphics Processing Units (GPUs) has revolutionized uncertainty estimation. NNs, with their ability to model complex, non-linear relationships, provide a flexible framework for uncertainty estimation. This is further enhanced by Mixed Density Networks (MDNs), which allow for the modeling of complex, multi-modal distributions. The parallel processing capabilities of GPUs expedite the training of these large, complex models, making it feasible to implement computationally intensive techniques such as Variational Inference and Kullback–Leibler (KL) divergence. UncertaintyPlayground leverages these advancements, offering Gaussian Processes (GPs) and MDNs for regression tasks. The library abstracts away the complexities of these techniques, allowing users to focus on their data and tasks. Designed to be user-friendly yet performant, it serves as a valuable tool for researchers and practitioners in machine learning, as well as less technical users and experts in statistics, economics, and bioinformatics.

The rest of the paper is organized as follows: Section 2 briefly covers the research question, necessary theoretical foundations, existing literature, and similar libraries for uncertainty estimation. Section 3 discusses the library's algorithm and workflow, including modules for model definition, training, and plotting prediction distributions. This section also discusses the parallelization capacity of the package . Section 4 describes the unit tests necessary for ensuring the build and outlines code maintenance strategies, including a local and an online workflow for continuous integration. Section 5 discusses the process of documenting the code and creating a website for the code documentation. Section 6, through an example, demonstrates the output of UncertaintyPlayground for the user. Section 7 provides additional ideas for further developing the library, and Section 8 concludes the paper.

\section{Research Question and Relevant Literature}
\subsection{Research Question}
This project's primary research question is how to estimate uncertainty in supervised learning tasks in an easy-to-use manner. The relevant literature includes works on GPs \cite{rasmussen2005gaussian} and MDNs \cite{bishop1994mixture}. The GP implementation is based on Sparse and Variational Gaussian Process techniques (SVGP) \cite{titsias2009variational}. We briefly highlight SVGP and why it was chosen for this library.

Neural Networks provide a flexible framework for modeling complex patterns in data. In particular, SVGPRs and MDNs, which are neural network-based implementations of Gaussian Processes and Mixture Models, offer a powerful approach to modeling uncertainty. SVGPRs allow for efficient and scalable Gaussian Process Regression, while MDNs enable the modeling of complex, multi-modal distributions. UncertaintyPlayground aims to provide these neural network-based techniques and tackle issues with these models' algorithmic complexity through methods such as Variational Inference and KL divergence.

\subsection{Relevant Literature}
\subsubsection{Gaussian Processes}
Gaussian Processes (GPs) have been recognized as a fundamental tool in machine learning, providing function priors for various tasks. In cases with Gaussian likelihoods, inference can be performed in a closed-form manner. However, for non-Gaussian likelihoods, posterior and marginal likelihood approximations are required \cite{kuss2005assessing, nickisch2008approximations}. Traditional GP regressions have a significant drawback: the computational cost of the exact calculation of the posterior and marginal likelihood scales as \(O(N^3)\) in time and \(O(N^2)\) in memory, where \(N\) is the number of training examples. 

Sparse and Variational Gaussian Process Regression (SVGPR) is a particular implementation of GPs that allows for efficient and scalable Gaussian Process Regression \cite{titsias2009variational,hensman2013gaussian}. SVGPR addresses this issue by choosing \(M\) inducing variables that are \(M \ll N\)  to summarize the entire posterior. This reduces the computational cost to \(O(N M^2 + M^3)\) in time and \(O(N M + M^2)\) in memory \cite{burt2019rates}. SVGPR works by minimizing the Kullback-Leibler (KL) divergence between the approximate and the true posterior, which allows us to learn the model parameters via gradient descent. This is significantly lower than the \(O(N^3)\) time complexity of standard Gaussian Process Regression, making SVGPR a more scalable alternative for large datasets \cite{hensman2013gaussian}.

\subsubsection{Mixed Density Networks}
Mixed Density Networks (MDNs) represent another powerful approach in machine learning, particularly for modeling complex, multi-modal distributions. MDNs combine the flexibility of neural networks with the robustness of mixture models, enabling them to capture intricate patterns in data \cite{bishop1994mixture}. 

The computational complexity of MDNs is primarily dependent on the number of mixture components and the dimensionality of the data. Unlike GPs, the complexity of MDNs does not increase cubically with the number of data points. However, as the number of mixture components or the dimensionality of the data increases, the computational cost of training an MDN can become substantial. Despite this, the parallel processing capabilities of modern hardware, such as Graphics Processing Units (GPUs), can significantly expedite the training of these models, making MDNs a feasible approach for large-scale, high-dimensional tasks.

\section{Code Base}

The methodology involves the implementation of SVGP and MDN using the GPyTorch \cite{gardner2018gpytorch} and PyTorch \cite{paszke2019pytorch} libraries. As mentioned, the GP implementation is based on SVGP, hence SVGPR, while MDN is implemented purely in PyTorch. Other dependencies for this project are Numpy \cite{numpy}, Scikit-learn \cite{scikit-learn}, Matplotlib \cite{matplotlib}, and Seaborn \cite{seaborn}. Details regarding the version of these packages can be found in the \texttt{README.md} and \texttt{requirements.txt} files.

The entire code base (version 0.1) has been depicted in \autoref{fig:project_structure}. The workflow of the package goes through \texttt{models}, \texttt{trainers} (supported via supplementary sub-modules in \texttt{utils}), and then the predictions are plotted via \texttt{predplot}. After the development of the library, several unit tests were designed to ensure its functionality and placed in the \texttt{tests} directory. Finally, the documentation and some examples can be found in \texttt{docs} and \texttt{examples} folders.

\begin{minipage}{\linewidth}
\begin{mdframed}
\dirtree{%
.1 \textit{UncertaintyPlayground/}.
.2 README.md.
.2 setup.py.
.2 LICENSE.
.2 MANIFEST.in.
.2 tox.ini.
.2 mkdocs.yml.
.2 docs/.
.3 bib.md.
.3 example.md.
.3 gen\_ref\_pages.py.
.3 index.md.
.3 README.md.
.2 examples/.
.3 compare\_models\_example.py.
.3 mdn\_example.py.
.3 svgp\_example.py.
.2 uncertaintyplayground/.
.3 requirements.txt.
.3 models/.
.4 mdn\_model.py.
.4 svgp\_model.py.
.3 trainers/.
.4 base\_trainer.py.
.4 mdn\_trainer.py.
.4 svgp\_trainer.py.
.3 predplot/.
.4 grid\_predplot.py.
.4 mdn\_predplot.py.
.4 svgp\_predplot.py.
.3 utils/.
.4 early\_stopping.py.
.4 generate\_data.py.
.3 tests/.
.4 test\_early\_stopping.py.
.4 test\_generate\_data.py.
.4 test\_mdn\_model.py.
.4 test\_mdn\_predplot.py.
.4 test\_mdn\_trainer.py.
.4 test\_svgp\_model.py.
.4 test\_svgp\_predplot.py.
.4 test\_svgp\_trainer.py.
}
\end{mdframed}

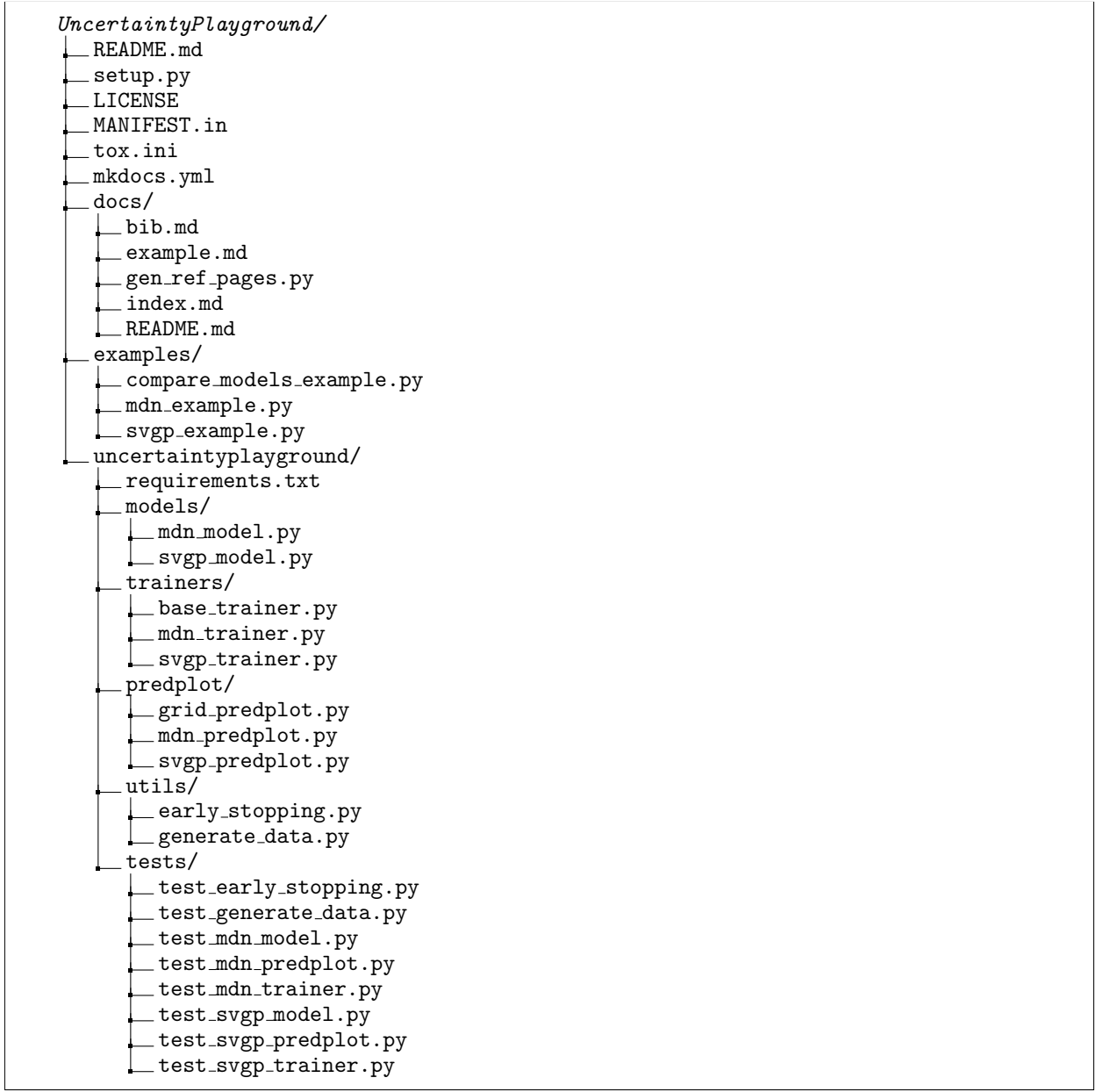
\captionof{figure}{Project structure of UncertaintyPlayground v0.1}
\label{fig:project_structure}
\end{minipage}

\subsection{Models}
As mentioned already, in the current version of the package, two models are implemented: an MDN and an SVGP, defined in the modules \texttt{mdn\_model} and \texttt{svgp\_model}, respectively.

The \texttt{MDN} model is designed to predict multi-modal distributions. It is particularly useful in scenarios where the data does not adhere to a simple Gaussian distribution. This model includes a neural network with one hidden layer comprised of 10 neurons and a Tanh (hyperbolic tangent) activation function. This network generates parameters for a mixture of Gaussians, which include mixture weights (\texttt{z\_pi}), means (\texttt{z\_mu}), and standard deviations (\texttt{z\_sigma}).

The model provides three options for prediction methods: `max\_weight\_mean', `max\_weight\_sample', and `average\_sample'. These approaches represent different strategies for choosing components from the Gaussian mixture and generating predictions:

\begin{itemize}
\item \textbf{Max Weight Mean}: This method selects the component with the highest weight and uses its mean as the prediction. It is a deterministic strategy and tends to produce the most probable prediction.
\item \textbf{Max Weight Sample}: This method selects a component from the mixture based on the weights, then samples from the selected Gaussian component to yield a prediction. This is a stochastic strategy, introducing variability into the predictions.
\item \textbf{Average Sample}: This method generates multiple samples from the mixture, each time selecting a component based on the weights and sampling from the selected Gaussian component. The final prediction is the average of these samples, providing a balance between the deterministic 'max\_weight\_mean' and the stochastic 'max\_weight\_sample.'
\end{itemize}

 The prediction method selection depends on the particular use case and the trade-off between variability and predictability in the predictions. In all cases, a loss function, \texttt{mdn\_loss}, computes the target variable's negative log-likelihood given the mixture's predicted parameters. The forward pass of the model computes these parameters based on the input, and a sample method is included to generate samples from the output distribution based on the input.

The \texttt{SVGP} model is a Sparse Variational Gaussian Process model designed for large-scale regression tasks where traditional Gaussian Process models are computationally infeasible. It employs a subset of data points, referred to as inducing points, to approximate the full Gaussian process. This approximation reduces the computational complexity from cubic, making the uncertainty estimation more scalable to large datasets. The \texttt{SVGP} model inherits from the \texttt{ApproximateGP} class provided by the \texttt{gpytorch} library. The model is initialized with a variational distribution strategy, defined using the provided inducing points and data type. The model is designed to learn the inducing locations during the training process. The \texttt{SVGP} model includes a constant mean module (\texttt{mean\_module}) and a scaled radial basis function (RBF) kernel (\texttt{covar\_module}). The forward pass of the model computes a multivariate normal distribution with a mean and covariance provided by the mean and covariance modules.

While both models compute the output distribution given input during the forward pass, they differ significantly in their representation of the output distribution. The \texttt{MDN} uses a mixture of Gaussians to represent complex, multi-modal distributions. In contrast, \texttt{SVGP} employs Gaussian processes, more specifically, a sparse approximation, suitable for modeling functions with smooth, continuous outputs in large-scale settings.

\subsection{Utilities}

The \texttt{utils} module in the project comprises two smaller sub-modules: \texttt{early\_stopping} and \texttt{generate\_data}. These sub-modules offer vital functionality to control the model training process and generate synthetic data for testing, respectively.

The sub-module \texttt{early\_stopping} encompasses the \texttt{EarlyStopping} class, which is designed to halt the training procedure of a model when a specified performance metric ceases to improve over several consecutive epochs. The extent of tolerance for non-improvement is decided by the \texttt{patience} parameter. Furthermore, a comparison function, \texttt{compare\_fn}, compares the metric values. By default, the function evaluates with a logical 'less than' comparison, implying that lower metric values are superior since we minimize the loss functions in both situations. In case of an improvement in the validation metric, the model's state is saved. This feature is crucial as it aids in the prevention of overfitting and reduces unnecessary computation.

The \texttt{generate\_data} sub-module provides a function, \texttt{generate\_multi\_modal\_data}, which generates multi-modal data helpful in testing the models. This function accepts the total number of data samples to be generated, along with a list of dictionaries specifying the modes of the distribution. Each dictionary represents a mode defined by its mean, standard deviation, and weight. Here, the weight dictates the proportion of total samples that will be drawn from the specific mode. The function returns a NumPy array of the generated data samples. Such functionality is pivotal for generating synthetic data that mimics the complex, multi-modal distributions the models aim to handle.

\subsection{Trainers}
The training aspect of the project is catered by a class \texttt{BaseTrainer} and the two model-specific trainers classes: \texttt{MDNTrainer} and \texttt{SVGPTrainer}. The base trainer provides the general functionality for training machine learning models, which is then specialized for training the MDN and SVGP models in their respective trainers. 

The \texttt{BaseTrainer} class provides a basic structure for the training procedure, defining common functionalities that are universally required for model training, such as setting up the training, test, and validation datasets via PyTorch DataLoaders, defining the loss function and optimizer, and setting up the training loop just to name a few arguments. This base class is designed to be extended by more specific trainer classes to train particular models, and the training process itself is implemented within the child classes. One important capability of this class is providing sample weights to have a weighted loss function. By default, all the weights are set to 1.

One child class of \texttt{BaseTrainer} is the \texttt{MDNTrainer}. This class is specialized for training the MDN model, as defined in the \texttt{mdn\_model} module. The trainer initializes the MDN model with a specified number of hidden neurons and Gaussian components and employs a specified optimizer for training. During training, the trainer iterates over the data batches, using the MDN's loss function to compute the training loss and update the model parameters. Additionally, the trainer calculates the validation loss after each epoch to track the model's performance on unseen data. It utilizes the \texttt{EarlyStopping} mechanism mentioned earlier to halt training if the validation loss does not improve over a specified number of epochs, which helps prevent overfitting.

The \texttt{MDNTrainer} also includes a method for making predictions with uncertainty, \texttt{predict\_with\_uncertainty}. This method feeds new data into the trained MDN model and returns the parameters of the predicted Gaussian mixture, including the mixture weights, means, and standard deviations, as well as samples from the predicted distribution.

The \texttt{SVGPTrainer} follows a similar structure and role as the \texttt{MDNTrainer}, providing functionality for training the SVGP model defined in the \texttt{svgp\_model} module. During the initialization of the \texttt{SparseGPTrainer}, the class takes in the number of inducing points as an argument and initializes the SVGP model with these inducing points. It also sets up a Gaussian likelihood for the model. The training method in this class contains a training loop that uses the VariationalELBO loss function, which is related to the KL divergence mentioned earlier. Similar to \texttt{MDNTrainer}, the method also includes early stopping functionality, checking the validation loss after each epoch and stopping the training if the validation loss does not improve for a set number of epochs.

The function \texttt{predict\_with\_uncertainty} varies between \texttt{SVGPTrainer} and \texttt{MDNTrainer} due to the nature of the models they handle. The trainer in \texttt{SVGPTrainer} also contains a \texttt{predict\_with\_uncertainty} method to make predictions and the associated uncertainty using the trained model. In \texttt{SVGPTrainer}, this method passes the input tensor through the trained SVGP model and its Gaussian likelihood, producing a Gaussian predictive distribution for each data point. The mean and variance of these distributions are then interpreted as the predicted output and its associated uncertainty. In contrast, \texttt{MDNTrainer} works with MDNs, yielding a Gaussian Mixture Model (GMM) as output. This results in multiple Gaussian components per data point (three per component), each contributing to the final prediction and its uncertainty. Consequently, despite both methods aiming to estimate prediction uncertainty, the structure of the output distributions and the procedure of extracting uncertainty information differ considerably.

\subsection{Prediction Plots}
The \texttt{predplot} module is dedicated to visualizing the models' predictions and the corresponding uncertainty. This module includes three sub-modules; \texttt{svgp\_predplot}, \texttt{mdn\_predplot} and \texttt{grid\_predplot}. As the names suggest, the first two sub-modules are for plotting the predictions with MDN and SVGP models for a single instance, while \texttt{grid\_predplot} takes as input any of the two functions and plots the same function for two or more instances. In all cases, Matplotlib is the main library for visualizations.

The \texttt{svgp\_predplot}, \texttt{mdn\_predplot} sub-modules each contain a \texttt{compare\_distributions\_} (svgpr or mdn) function which takes a trained MDN model, an input instance, and optionally the actual outcome(s), and plots the predicted distribution for that given instance, and compare the final scalar prediction against the actual outcome value (given that the latter is provided). The predicted distribution is shown as a Kernel Density Estimate (KDE) with the Seaborn library, and the predicted value and actual value(s) are represented as vertical lines on the plot. 

The two functions were separated for several reasons. For the MDN model, the dedicated function plots the learned GMM for each test point, indicating (and printing) the predicted means, variances, and weights of the Gaussian components. For the SVGP model, the function plots the predicted mean and variance at each test point. As a result, there are many more possibilities for the MDN techniques regarding how the different Gaussian distributions can be represented (for instance, separately or KDE as it currently stands).

The \texttt{grid\_predplot} sub-modules contains a function and a class: \texttt{plot\_results\_grid} and \texttt{DisablePlotDisplay}. The \texttt{plot\_results\_grid} method generates a grid of scatter plots for given data sets. Each subplot presents the ground truth data as a scatter plot while superimposing the model's prediction and the associated uncertainty as a shaded area. This method is generic and can be used for both models. Additionally, this sub-module contains a \texttt{DisablePlotDisplay} class, a special utility class included in the codebase to control the display of plots during the execution of the script. This class is especially useful when running the code in an environment where the graphical display of plots is not possible or not desirable, such as with our automated tests for the plots.

\subsection{Parallelization}
The UncertaintyPlayground codebase effectively leverages parallelization at various stages to speed up the computations. PyTorch's native support for GPU acceleration is used to parallelize computations involved in model training. This functionality is implemented within the \texttt{BaseTrainer} class, where the \texttt{device} attribute determines whether computations are to be performed on a CPU or GPU.

In addition to GPU acceleration, the software utilizes parallelization when loading data, significantly reducing the data-loading time, especially for large datasets. The data-loading step in PyTorch can be easily parallelized across multiple CPUs using the \texttt{num\_workers} parameter of the \texttt{DataLoader} class. By setting \texttt{num\_workers} to be greater than zero, the data loading tasks are divided among multiple subprocesses. This approach allows to load data in parallel and ensures that the GPU is not idle. In contrast, the data is being loaded, thus maximizing the GPU utilization and overall performance.

While currently, the project does not explicitly support multi-GPU configurations for model training; this could be added in the future using PyTorch's DataParallel wrapper for distributed computation. This could further speed up model training, especially for large models and datasets. However, this enhancement would require careful management of memory and synchronization of model parameters across the different GPUs, which would be a complex task warranting further investigation.

\section{Code Maintenance}
The codebase is shared and maintained using GitHub, which facilitates collaborative development. The repository includes a \texttt{.gitignore} file, containing specific files and directories that Git should ignore, such as the package build files, logs, and local configurations that are not intended to be part of the shared codebase. Additionally, we ignore the files that are generated when running local integration with Tox and Pyenv.

Unit tests are implemented using Python's built-in \texttt{unittest} framework. There are eight unit tests in total, ensuring the correctness and robustness of the codebase. The use of unit tests also contributes to the modularity of the code, as each component can be tested independently.

Continuous Integration (CI) was designed to work both locally (offline) and online. The CI system automatically tests the code with multiple Python versions (3.8, 3.9, and 3.10), ensuring that the code works as expected across different platforms and Python versions. Local integration is implemented with Tox (\texttt{tox.ini}) and Pyenv, and online integration is set up through GitHub Workflows (\texttt{.github/workflows/ci\_cd.yml}). The dual implementation of CI is due to the advantages and disadvantages of these techniques. On the one hand, GitHub Workflow integration tests compatibility on Mac, Windows, and Linux OS, while Tox is specific to the local OS. On the other hand, GitHub workflows have limited testing calls for private repositories (lifted for public projects), while Tox can be used free of charge.

\section{Code Documentation}
The code documentation is based on Google-style docstrings, which are used throughout the code to describe the purpose and behavior of classes, methods, and functions. The MkDocs tool, a fast and simple static site generator, produces the project documentation from these docstrings and additional markdown files. The docstrings themselves are converted via the MkDocStrings software extension. The markdown files used for MkDocs are located in the \texttt{docs} directory, and the \texttt{mkdocs.yml} configuration file is placed at the project's root directory. This approach was specifically chosen to ensure that the documentation always stays in sync with the code.

\section{Usage}
In this section, we discuss the output for the users and some capabilities of the package. We shed light on the capability of this data with a real data set commonly used in ML, namely the California housing data. This dataset provides a realistic scenario for the models. However, the outcome variable, the average house value, is not necessarily multi-modal. In the \texttt{examples} folder of the package and the `Usage' section of the documentation, you find a better example for MDN where simulated data which contain multi-modal distributions can be used to test specific aspects of the MDN models.

First, we load California housing data and convert it to the desired floating-point format as shown in \autoref{fig:california-housing}. Then, we initialize and train an SVGP model with 100 inducing points as depicted in \autoref{fig:california-svgp-code} and obtain the output for the training. It can be observed that the model starts learning and then stops at the 30\textsuperscript{th} epoch. Please note here that we do not set a seed for Numpy and PyTorch, but when applying this library, it is good practice to set the two seeds.

\begin{figure}[htbp]
    \begin{lstlisting}[frame=single]
>>> from sklearn.datasets import fetch_california_housing
>>> from sklearn.model_selection import train_test_split
>>> import numpy as np
>>> california = fetch_california_housing()
>>> X = np.array(California.data, dtype=np.float32)
>>> y = np.array(California.target, dtype=np.float32)
>>> X_train, X_test, y_train, y_test = train_test_split(X, y, test_size=0.2, random_state=42)
\end{lstlisting}
\caption{Loading the California Housing dataset}
\label{fig:california-housing}
\end{figure}

\begin{figure}[htbp]
    \begin{lstlisting}[frame=single]
>>> from uncertaintyplayground.trainers.svgp_trainer import SparseGPTrainer
>>> california_trainer_svgp = SparseGPTrainer(
        X=X_train, 
        y=y_train, 
        num_inducing_points=100, 
        num_epochs=30, 
        batch_size=512, 
        lr=0.1, 
        patience=3)
>>> california_trainer_svgp.train()
Epoch 1/30, Weighted Loss: 1.731, Val MSE: 1.380385, Val R2: -0.000
Epoch 2/30, Weighted Loss: 1.676, Val MSE: 1.381376, Val R2: -0.001
...
Epoch 30/30, Weighted Loss: 1.514, Val MSE: 0.940440, Val R2: 0.319
    \end{lstlisting}
    \caption{Initializing and Training SVGPR}
    \label{fig:california-svgp-code}
\end{figure}    

We can apply the same kind of pipeline for the MDN model as illustrated in 
\autoref{fig:california-mdn-code}. Aside from the function arguments, the outputs differ from those of \autoref{fig:california-svgp-code}. The first difference is that other outputs are not given aside from the loss metric. It may be argued that inference with multiple Gaussian distributions at every epoch may be undesirable. However, this feature will be added in the next versions of the package. The second difference is that the training stopped earlier than expected. This is due to the \texttt{patience} argument, which triggers the early stopping at the default value of 10, meaning that validation loss did not improve after the 11\textsuperscript{th} epoch for another ten epochs. Hence the training was stopped, and the best model was returned.

\begin{figure}[htbp]
    \begin{lstlisting}[frame=single]
>>> from uncertaintyplayground.trainers.mdn_trainer import MDNTrainer
>>> california_trainer_mdn = MDNTrainer(
        X=X_train, 
        y=y_train, 
        num_epochs=100, 
        lr=0.001, 
        dense1_units=50, 
        n_gaussians=10)
>>> california_trainer_mdn.train()
Epoch 1/100, Training Loss: 1.450, Validation Loss: 1.522
Epoch 2/100, Training Loss: 1.382, Validation Loss: 1.458
...
Epoch 11/100, Training Loss: 1.381, Validation Loss: 1.261
...
Epoch 21/100, Training Loss: 1.412, Validation Loss: 1.293
Early stopping after 21 epochs
    \end{lstlisting}
    \caption{Initializing and Training MDN}
    \label{fig:california-mdn-code}
\end{figure}    

We must note that we do not show an example of \texttt{predict\_with\_uncertainty} in this paper for brevity. These functions are also applied for generating prediction plots and will be discussed briefly. Please note that this function for both models produces the learned parameters of the model. More information can be found on this topic in the documentation.

After the models have been trained, we visualize the prediction intervals. For the SVGP model, use the function \texttt{compare\_distributions\_svgpr} and the function \texttt{plot\_results\_grid}.

\begin{figure}[!h]
\centering
\includegraphics[width=0.5\columnwidth]{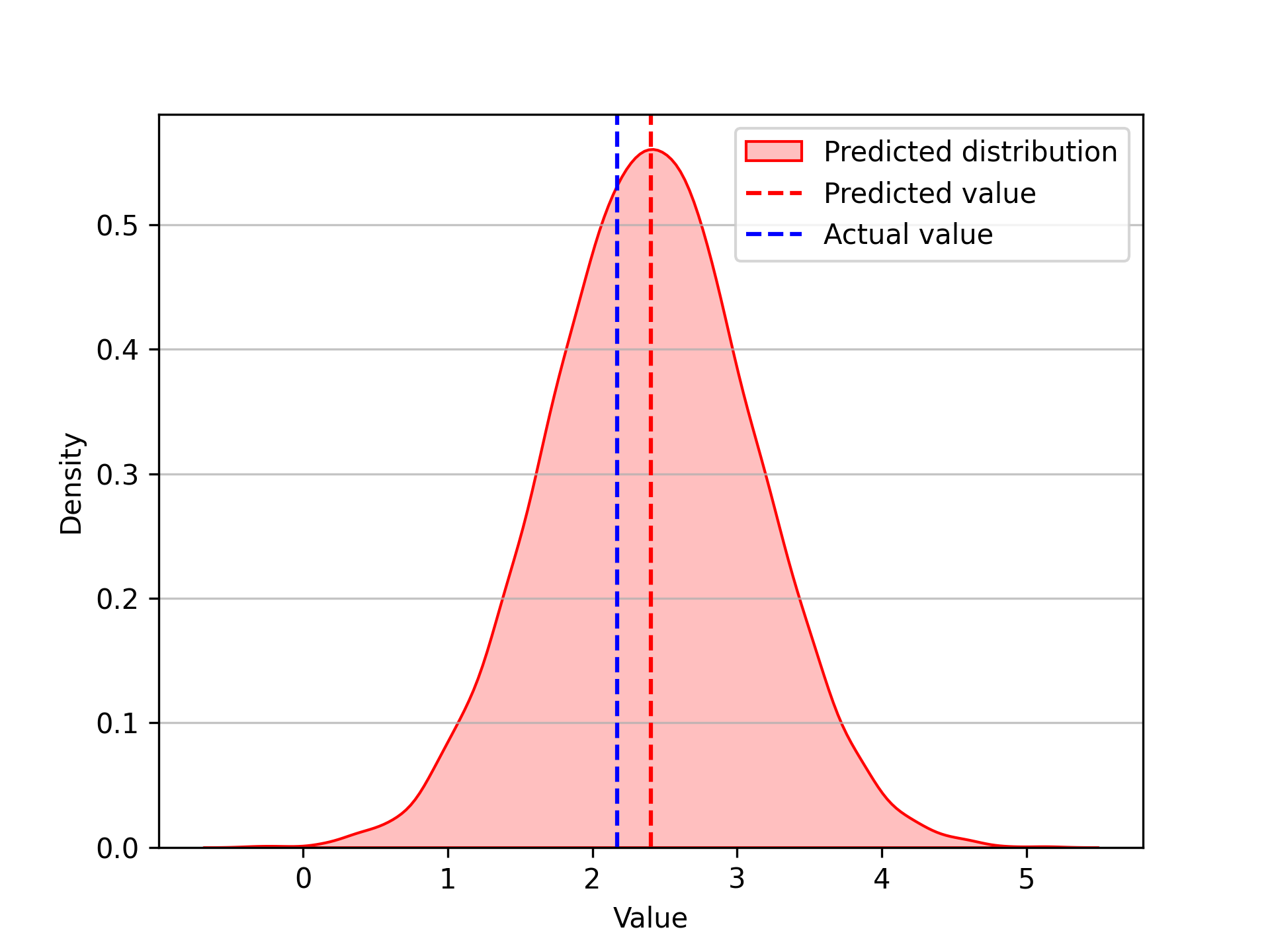}
\caption{SVGPR \texttt{compare\_distributions\_svgpr} for instance number 900}
\label{fig:california-plot-svgp-solo}
\end{figure}

\begin{figure}[!h]
\centering
\includegraphics[width=0.8\columnwidth]{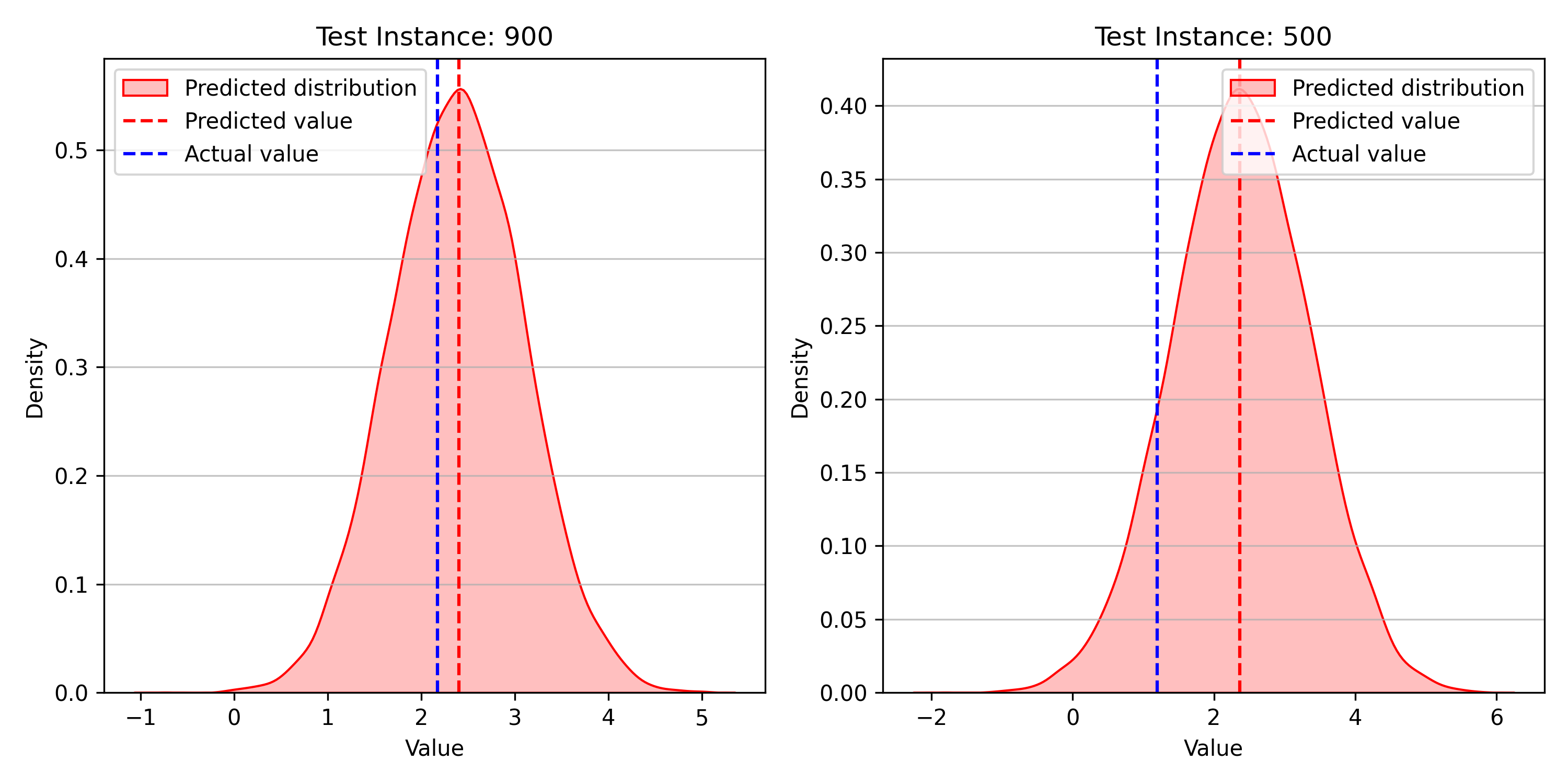}
\caption{SVGPR \texttt{plot\_results\_grid} on two instances}
\label{fig:california-plot-svgp-grid}
\end{figure}

\begin{figure}[!h]
\centering
\includegraphics[width=0.5\columnwidth]{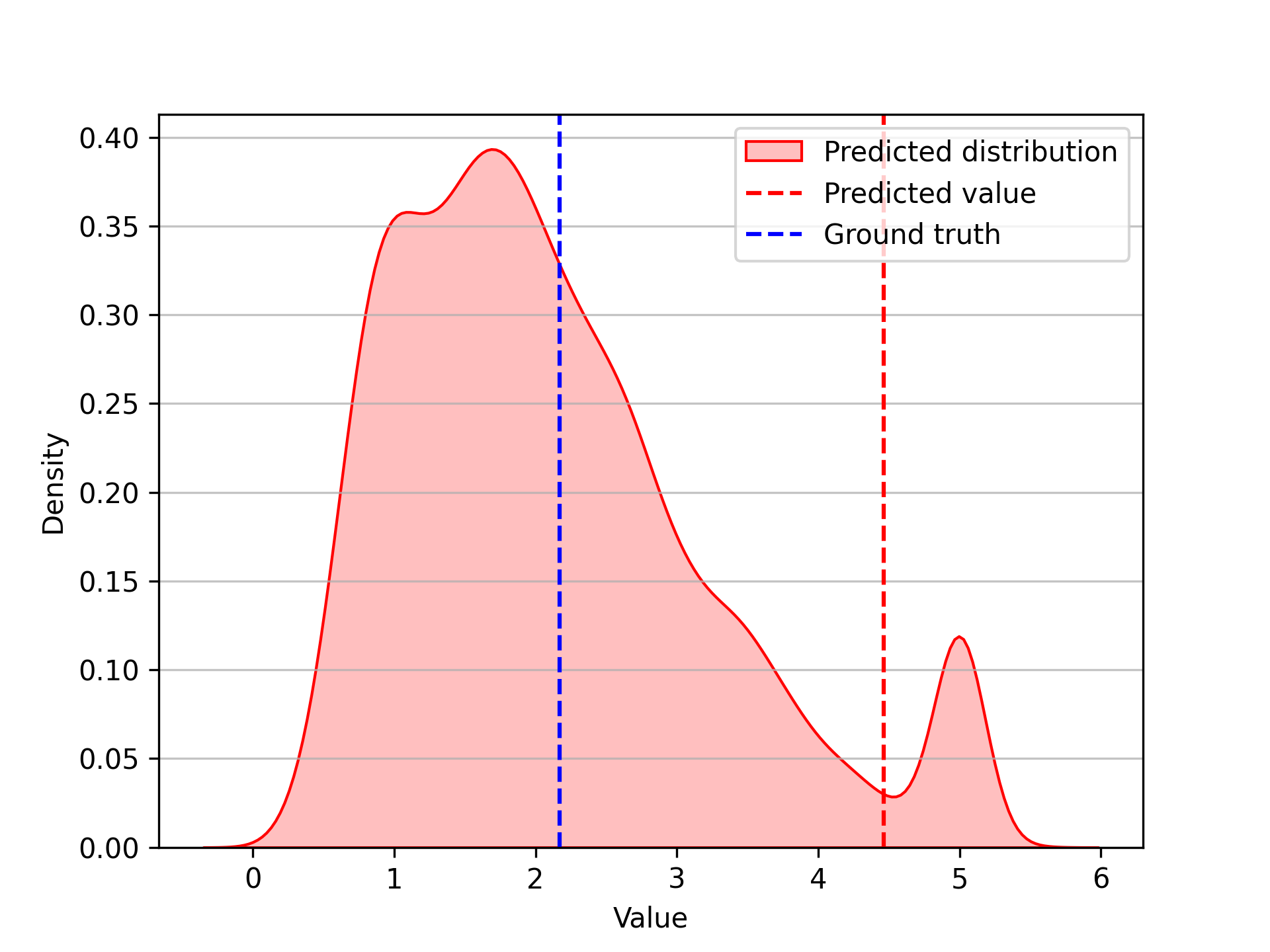}
\caption{MDN \texttt{compare\_distributions\_mdn} for instance number 900}
\label{fig:california-plot-mdn-solo}
\end{figure}

\begin{figure}[!h]
\centering
\includegraphics[width=0.4\columnwidth]{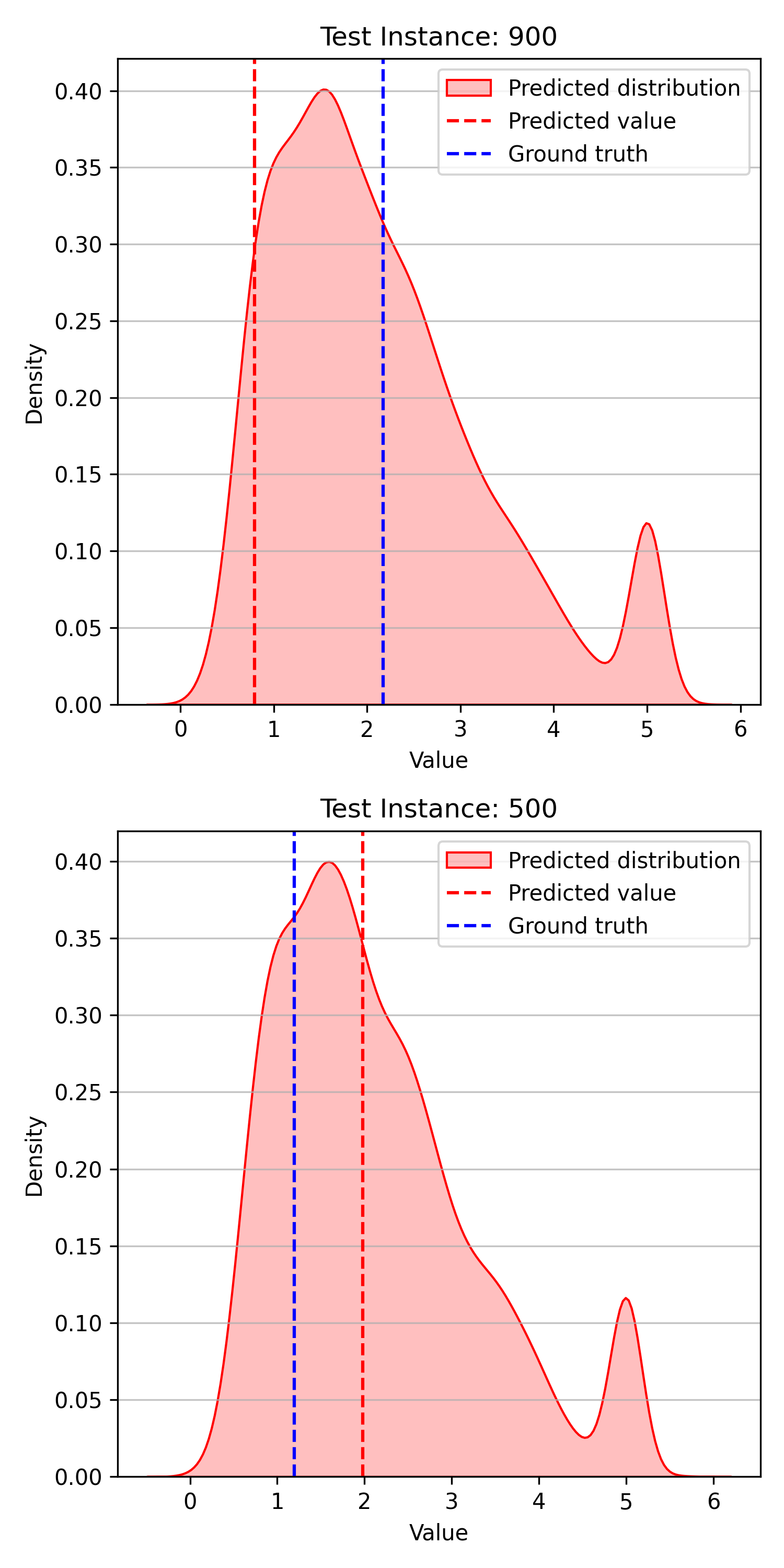}
\caption{MDN \texttt{plot\_results\_grid} on two instances}
\label{fig:california-plot-mdn-grid}
\end{figure}

Next, we visualize the predictions of the MDN model using the function \texttt{compare\_distributions\_mdn} and the function \texttt{plot\_results\_grid}. These workflows can be found in \autoref{fig:california-plot-code}. The four produced plots are shown in \autoref{fig:california-plot-svgp-solo}, \autoref{fig:california-plot-svgp-grid} and \autoref{fig:california-plot-mdn-solo} and \autoref{fig:california-plot-mdn-grid}. 

\begin{figure}[htbp]
\begin{lstlisting}[frame=single]
>>> from uncertaintyplayground.predplot.mdn_predplot import compare_distributions_mdn
>>> from uncertaintyplayground.predplot.svgp_predplot import compare_distributions_svgpr
>>> from uncertaintyplayground.predplot.grid_predplot import plot_results_grid
>>> compare_distributions_svgpr(
        trainer=california_trainer_svgp, 
        x_instance=X_test[900, :], 
        y_actual=y_test[900])
>>> plot_results_grid(
        trainer=california_trainer_svgp, 
        compare_func=compare_distributions_svgpr,
        X_test=X_test, 
        Y_test=y_test, 
        indices=[900, 500], 
        ncols=2)
>>> compare_distributions_mdn(
        trainer=california_trainer_mdn, 
        x_instance=X_test[900, :], 
        y_actual=y_test[900])
>>> plot_results_grid(
        trainer=california_trainer_mdn, 
        compare_func=compare_distributions_mdn, 
        X_test=X_test, 
        Y_test=y_test, 
        indices=[900, 500], 
        ncols=1)
\end{lstlisting}
\caption{Plotting results from both SVGPR and MDN models}
\label{fig:california-plot-code}
\end{figure}    

The plots and their predictive distributions illustrated how to diagnose the model's performance and better understand the underlying data structure. We do not discuss the quality of the model in this report. Nonetheless, it is helpful to see that the MDN model may find skewed data that SVGPR does not capture. Whether this model is better than SVGPR needs proper metrics and validation and is beyond the scope of this report.

\section{Further Development}
The UncertaintyPlayground project, as a dynamic platform for uncertainty estimation, has numerous avenues for future enhancements and expansions. The opportunities for improvements are distributed across different elements of the package:

\begin{enumerate}

\item \textbf{Flexible Model Architectures:} For the MDN model, it would be beneficial to make the neural network architecture modular, allowing for the incorporation of different layers and activation functions. For the SVGP model, adding the option to use different kernel functions could extend the model's flexibility.

\item \textbf{Improved Noise Modelling:} Introducing the capability to use different types of noise in the MDN model could significantly improve the quality of uncertainty estimates.

\item \textbf{Classification Capabilities:} Both the MDN and SVGP models could be extended to support binary and multi-class classification. This would involve modifying the likelihood function and the performance metric. This would, however, require extensive theoretical and empirical validation as MDNs and SVGPRs are not traditionally used for classification tasks \cite{hensman2015scalable}.

\item \textbf{Hardware Utilization:} The package could benefit from implementing multi-GPU support, which would allow for more efficient training of large models on large datasets. Optimizing the parallel data loading process for maximized CPU utilization could significantly improve overall performance.

\item \textbf{Improved Code Documentation:} The addition of type hints to the docstrings would offer better clarity and type checking, enhancing the readability and maintainability of the codebase. Additionally, the documentation can benefit from more examples.

\item \textbf{Benchmarking Performance:} In the further iterations of this package, the performance, both in terms of speed and accuracy of prediction, can be measured against other models. For instance, one can compare our approach with a traditional GPR for larger and smaller datasets since, as already discussed, GPR has an algorithmic complexity of \(O(N^3)\) and does not scale well beyond a few hundred observations.

\end{enumerate}

\section{Conclusion}
This paper comprehensively reviews a newly developed Python package, UncertainityPlayground, a library built for simplified uncertainty estimation with PyTorch and GPyTorch. The two machine learning algorithms, MDN and SVGP models, are implemented for fast and easy uncertainty estimation of continuous outcomes. The results demonstrate that both models can model complex data distributions and provide meaningful uncertainty estimates. The built-in plotting functionalities allow users to study the inferred distribution of given outcomes. The codebase is well-maintained, well-documented, and designed for future development.

\end{document}